%% file: draft_arXiv.tex
\documentclass[10pt,twocolumn,letterpaper]{article}

\usepackage{cvpr}
\usepackage{times}
\usepackage{epsfig}
\usepackage{graphicx}
\usepackage{amsmath}
\usepackage{amssymb}
\usepackage{multirow}
\usepackage{bm}
\usepackage[breaklinks=true,bookmarks=false]{hyperref}

\def\y{{\bm{y}}}
\def\Y{{\mathcal{Y}}}
\def\b{{\mathbf{b}}}
\def\B{{\mathcal{B}}}
\def\btheta{{\bm{\theta}}}
\def\bphi{{\bm{\phi}}}
\def\Image{{\bm{I}}}
\def\h{{\bm{h}}}
\def\x{{\bm{x}}}
\def\z{{\bm{z}}}
\def\W{{\bm{W}}}
\def\U{{\bm{U}}}
\DeclareMathOperator\relu{relu}

\cvprfinalcopy %

\def\cvprPaperID{****} %

\DeclareMathOperator*{\argmax}{\arg\!\max}

\begin{document}

\title{Sequential Person Recognition in Photo Albums with a Recurrent Network\thanks{This work was in part supported
    by Data61 (formerly National ICT Australia).
    All correspondence should be addressed to C. Shen
    (e-mail: {\tt chunhua.shen@adelaide.edu.au}).
  }
}

\author{
  Yao Li,
  Guosheng Lin,
  Bohan Zhuang,
  Lingqiao Liu,
  Chunhua Shen,
  Anton van den Hengel
 \\
 The University of Adelaide, Australia
}

\maketitle
\begin{abstract}
Recognizing the identities of people in everyday photos is still a very challenging problem for machine vision, due to non-frontal faces, changes in clothing, location, lighting and similar.
Recent studies have shown that rich relational information between
people in the same photo can help in recognizing their identities.
In this work, we propose to model the relational information between
people as a sequence prediction task.
At the core of our work is a novel recurrent network architecture,
in which relational information between instances' labels and appearance are modeled jointly.
In addition to relational cues, scene context is incorporated in our sequence prediction model with no additional cost.
In this sense, our approach is a unified framework for modeling both contextual cues and visual appearance of person instances.
Our model is trained end-to-end with a sequence of annotated instances in a photo as inputs, and a sequence of corresponding labels as targets.
We demonstrate that this simple but elegant formulation achieves
state-of-the-art performance on the newly released {People In Photo Albums (PIPA)} dataset.
\end{abstract}

\tableofcontents
\clearpage

\section{Introduction}
\label{sec:intro}
\input{intro.tex}
\input{related.tex}

\input{model.tex}

\section{Experiments}
\label{sec:experiments}
\input{experiment.tex}
\section{Conclusion}
\label{sec:conclusion}
\input{conclusion.tex}
{
\bibliographystyle{ieee}
\bibliography{egbib_2}
}

\end{document}

%% file: intro.tex
With the widespread adoption of digital cameras, the number of photos being taken has increased astronomically. 
The culture surrounding the use of these cameras means that a large proportion of these photos contain people.
The overwhelming volume of these images is creating a demand for smart tools to organize photos containing people. 
One cornerstone step is to recognize each person in these everyday images. 
Previous work~\cite{DBLP:conf/cvpr/AnguelovLGS07,DBLP:conf/cvpr/GallagherC08, DBLP:conf/cvpr/GallagherC09,DBLP:conf/eccv/WangGLF10,
DBLP:conf/cvpr/ZhangPTFB15,DBLP:conf/iccv/OhBFS15,
DBLP:conf/cvpr/LiBLSH16,DBLP:conf/eccv/OhBFS16} has shown that person 
recognition in such unconstrained settings remains a challenge problem
for machine vision due to various factors, such as non-frontal faces, varying lighting and settings, and even just the variability in the appearance of a face over time.

To tackle these challenges, in addition to the appearance of the face,  recent studies
~\cite{DBLP:conf/cvpr/AnguelovLGS07,DBLP:conf/cvpr/ZhangPTFB15,
DBLP:conf/iccv/OhBFS15,DBLP:conf/cvpr/LiBLSH16} have shown that contextual cues can help in recognizing peoples' identities in everyday photos. 
Other features of the individual, 
such as clothing~\cite{DBLP:conf/cvpr/AnguelovLGS07,DBLP:conf/cvpr/GallagherC08},
may also provide valuable cues.
The relationships between the person to be recognized and others, 
can also be a vital cue, however~\cite{DBLP:conf/eccv/WangGLF10,DBLP:conf/wacv/DaiCSH15,DBLP:conf/cvpr/LiBLSH16}.
To take advantage of different relation cues, probabilistic graphical models have been widely exploited~\cite{DBLP:conf/cvpr/AnguelovLGS07,
DBLP:conf/eccv/WangGLF10,DBLP:conf/wacv/DaiCSH15,
DBLP:conf/cvpr/LiBLSH16}.

In this work, we propose to model the rich relations between 
people in an image as a sequence prediction task. 
This is much motivated from the success of sequence prediction formulations in modeling relations between words in language problems
~\cite{DBLP:conf/nips/SutskeverVL14,DBLP:conf/cvpr/VinyalsTBE15}. 

\begin{figure}[t]
\begin{center}
\includegraphics[width=1\linewidth]{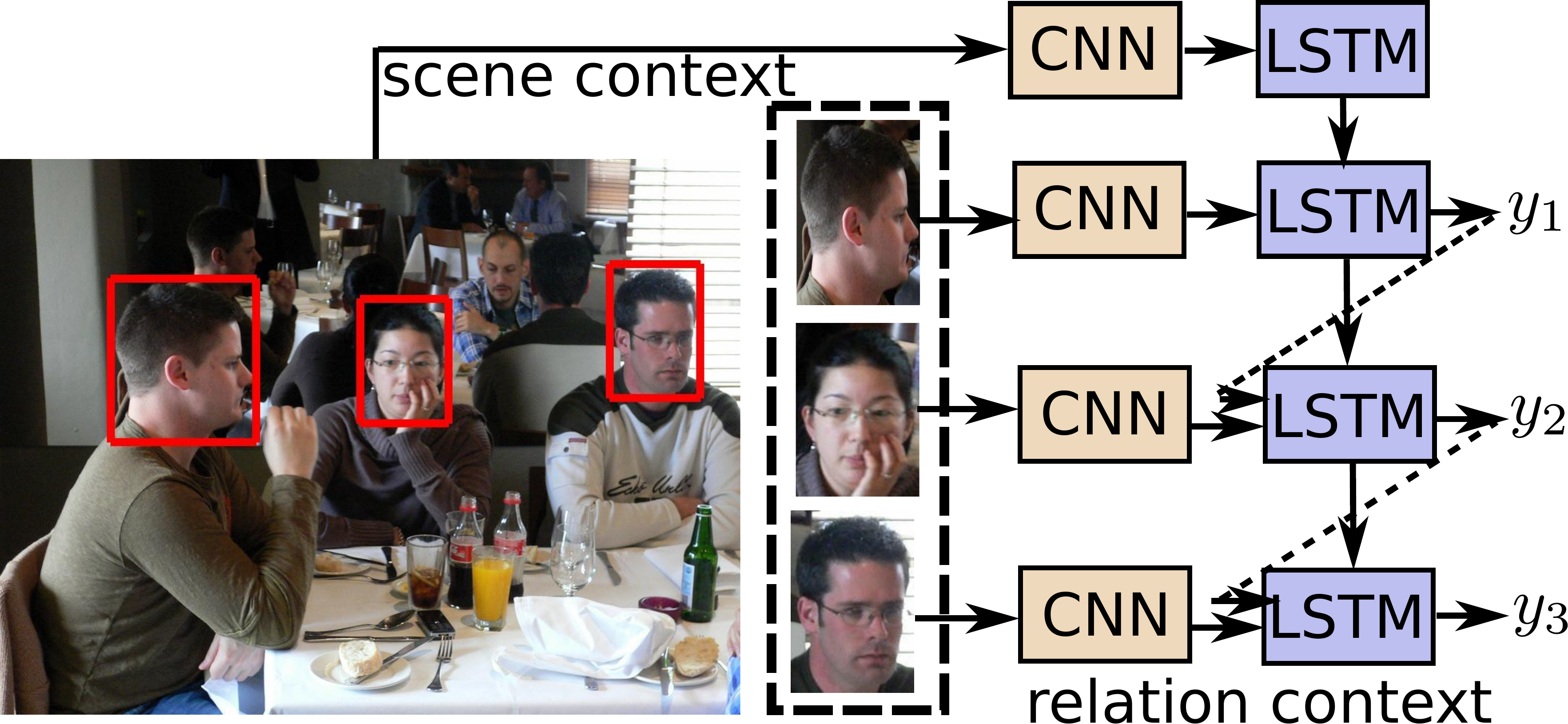} \\
\end{center}
\caption{Our approach performs person recognition in photo albums as a sequence prediction task.  Both contextual cues and instances' visual appearance are exploited in a unified framework.
}
\label{fig:overview} %
\end{figure}

In our work, we propose a novel Recurrent Neural Network (RNN) architecture for 
the sequence prediction task (see Fig.~\ref{fig:overview}), which consists of a Convolutional Neural Network (CNN) followed
a Long Short Term Memory
(LSTM) network~\cite{DBLP:journals/neco/HochreiterS97} . 
The LSTM has shown impressive performance on several sequence prediction problems
such as image captioning~\cite{DBLP:conf/cvpr/VinyalsTBE15}, video description~\cite{DBLP:conf/cvpr/DonahueHGRVDS15,
DBLP:conf/iccv/VenugopalanRDMD15}, multi-label image classification~\cite{DBLP:conf/cvpr/WangYMHHX16}, machine translation~\cite{DBLP:conf/nips/SutskeverVL14}, \etc. 
Initially
we feed the LSTM with global image information provided by the CNN.
At each subsequent step, the input to the LSTM is a joint embedding 
of the CNN feature representation of the current person instance and its predicted label at the last step. 
The LSTM then predicts the identity label for this person instance.
In this sense, the number of steps in our sequence model is of variable length, depending on the number 
annotated instances in the image.

\begin{figure}[t]
\begin{center}
\includegraphics[width=0.9\linewidth]{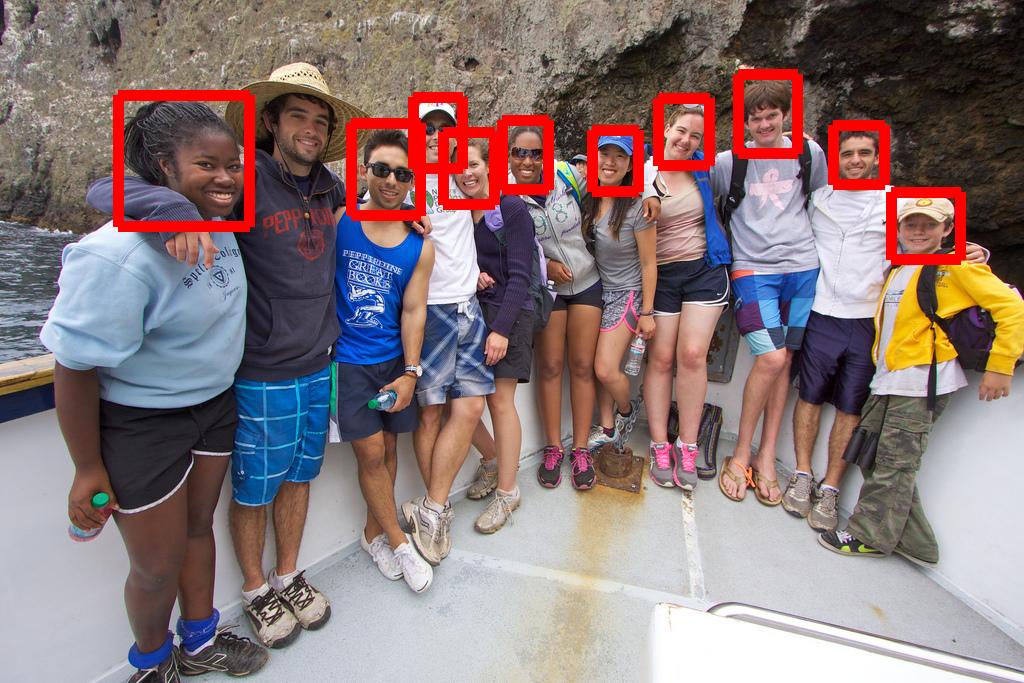} \\
\end{center}
\caption{In addition to each individual's appearance, there are  contextual cues which can help recognizing people in the photo, such as the relation context (\eg, which people are likely to appear together) and scene context (\eg, what is this place). 
}
\label{fig:example} %
\end{figure}

The two sources of contextual cues exploited in our model are the relation context and scene context (see Fig.~\ref{fig:example}). 
The relation context refers to the relational information between 
multiple people in the same image (\eg, some people are likely to appear together), which is naturally incorporated by our sequence prediction formulation.
Based on assumption that some people are more likely to appear in certain scenes, the scene can be used as a prior to indicate which identities tend to appear.
This cue is exploited in our model by feeding the global 
image feature to our sequence prediction model at the initial step, 
which informs the system about the scene content. 
As we will show in the experiments, both contextual cues are 
critical to the methods ability to achieve state-of-the-art performance.

To our knowledge, this is the first approach to formulate person recognition in photo albums as a sequence prediction task. 
This simple but elegant approach (a) enables modeling visual appearance and contextual cues in the same framework, (b) handles 
a variable number of instances in an image, (c) is end-to-end trainable. 
We demonstrate that our model achieves state-of-the-art performance 
on the \emph{People In Photo Albums (PIPA)} dataset~\cite{DBLP:conf/cvpr/ZhangPTFB15}, the benchmark 
dataset for person recognition in photo albums.

%% file: related.tex
\section{Related Work}
\label{sec:related}

\noindent\textbf{Person recognition in photo albums.} 
Person recognition in photo albums~\cite{DBLP:conf/cvpr/AnguelovLGS07,DBLP:conf/cvpr/GallagherC08,
DBLP:conf/eccv/WangGLF10,DBLP:conf/cvpr/ZhangPTFB15,
DBLP:conf/iccv/OhBFS15,DBLP:conf/cvpr/LiBLSH16} aims to recognize the identities of people in everyday photos. 
Intuitively, face region can be an important cue for the task, however, it may not entirely reliable, as in this unconstrained setting, people can have non-frontal, or even back-views.
This makes the problem much harder the classical face recognition. 

Recent studies on this topic have been boosted by the introduction of the PIPA dataset~\cite{DBLP:conf/cvpr/ZhangPTFB15}. 
In their original paper~\cite{DBLP:conf/cvpr/ZhangPTFB15}, the authors
proposed a method which combines information from three sources, including the full body, poselets~\cite{DBLP:conf/iccv/BourdevM09} and the DeepFace~\cite{DBLP:conf/cvpr/TaigmanYRW14} model.
Oh~\etal~\cite{DBLP:conf/iccv/OhBFS15} evaluated the importance of different cues for the task, such as different body regions, scene and human attributes. 
Li \etal~\cite{DBLP:conf/cvpr/LiBLSH16} proposed to incorporate contextual cues to the task, including group-level context and person-level context. 
However, the contextual models of~\cite{DBLP:conf/cvpr/LiBLSH16} 
are treated as the post-processing steps after the classification result, 
we instead exploit contextual information and visual cues in a unified 
framework.
This is particularly significant for the problem at hand because there are many possible identities for each detection, and non-frontal face cues can be extremely inconclusive.  A unified framework means that the cues are used collectively to exploit all of the available information, and will succeed over the greedy approach when the face recognition result is ambiguous. 

Our approach 
is also related to work on 
identifying 
people in group
photos~\cite{DBLP:conf/cvpr/GallagherC09,DBLP:conf/eccv/WangGLF10,
DBLP:conf/wacv/DaiCSH15, DBLP:conf/cvpr/MathialaganGB15}, as our LSTM framework naturally handles multiple people in the same photo. 
\\

\noindent\textbf{Modeling dependencies with RNNs. }
RNNs, and particularly LSTMs, have enjoyed much popularity recently in sequence modeling problems, largely due to their ability to model dependencies within sequences. 
For instance, LSTMs have been widely used in vision-to-language problems, such as image captioning~\cite{DBLP:conf/cvpr/VinyalsTBE15,wu2015image}, video description~\cite{DBLP:conf/cvpr/DonahueHGRVDS15,
DBLP:conf/iccv/VenugopalanRDMD15}, and visual question answering~\cite{wu2015ask,wu2016imagepami,wang2016fvqa}.

In comparison with the sequence prediction model in machine translation~\cite{DBLP:conf/nips/SutskeverVL14}, which contains both an encoder LSTM and decoder LSTM, we model both visual features and contextual cues using a single LSTM. 
In this sense, our model is closer to those used in image captioning~\cite{DBLP:conf/cvpr/VinyalsTBE15} as the LSTM output is sent to 
a classification layer at each step (except the initial step), as is the case in image captioning. 
However, the main difference with respect to image captioning models is that 
we have a visual feature input to the LSTM at every step (not only the initial step).

Although the identities of a group of people in a photo are better be described as a set than a sequence, there is a an obvious dependence between them.  It is this dependence that we seek to capture here using an RNN.  Despite being very popular for the task, the RNN model is not  sequence-specific, but rather, it can be employed to model a sequence by feeding the previous element of the sequence in as input.  What the RNN actually generates is an output conditioned on its internal state and input.  The output can be interpreted as a sequence, but it can equally be considered a set, or a variety of other types.  Wang \etal~\cite{DBLP:conf/cvpr/WangYMHHX16}, for instance, exploit LSTMs to model the dependencies between the multiple tags that different users apply to the same image (despite their being no natural order to the labels). Stewart \etal~\cite{stewart2015end} similarly use an RNN to model the dependencies between detections in an image, which also have no natural order.

\begin{figure*}[t]
\begin{center}
\includegraphics[width=1\linewidth]{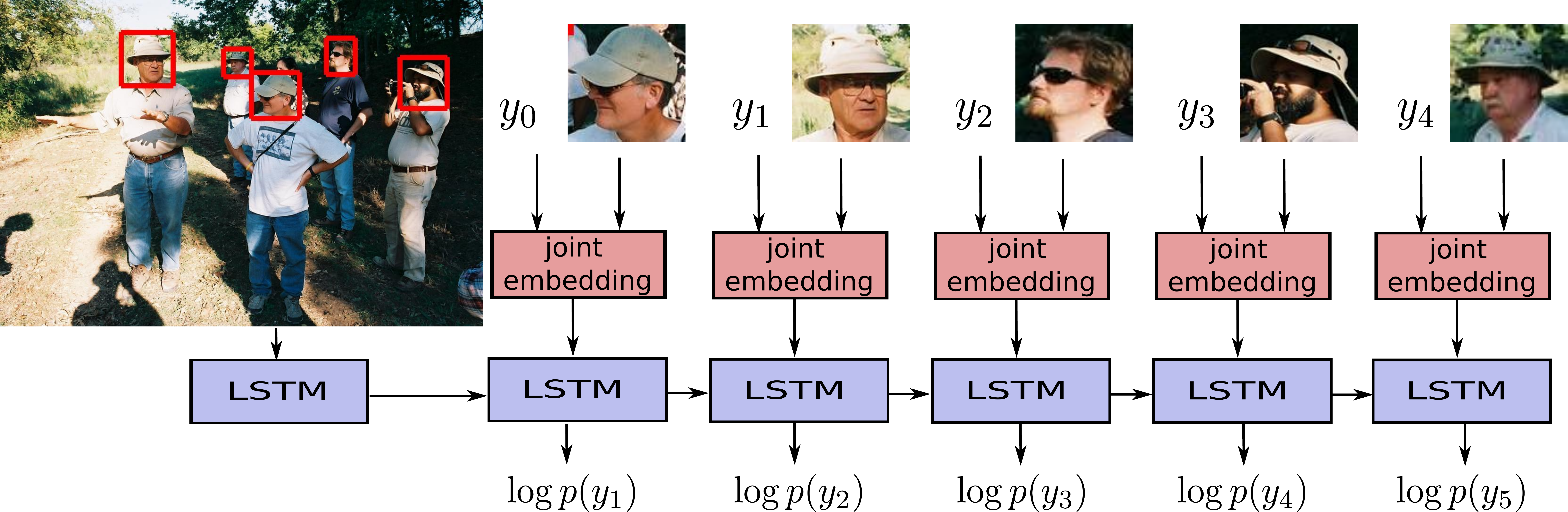} \\
\end{center}
\caption{Our sequence prediction approach for recognizing people in photo albums. For an image which may contain multiple people, 
our approach predicts the identity of each person in a sequence using an LSTM-based framework.
The initial state of the LSTM is informed by the 
scene context, and in each of the subsequent steps, 
the input to the LSTM is the joint embedding of the label of the last step and visual feature of the current instance (see Fig.~\ref{fig:joint_embed} for details).  The task of the LSTM is then to predict label of the current instance.
In this way, the relationships between people are naturally 
incorporated in our framework. 
Note that $y_0$ is the label of an auxiliary identity. 
}
\label{fig:pipeline} %
\end{figure*}

%% file: model.tex
\section{Model}
\label{sec:model}

We propose a sequence prediction approach for recognizing people 
in photo albums. 
As depicted in Fig.~\ref{fig:pipeline}, at each 
 step (except the first one), we jointly embed 
the previous label and the current image of an identity, 
which is then served as the input to an LSTM net. 
The LSTM not only tries to predict the current correct label, 
but also models the contextual information between people implicitly.
Our work is largely motivated by the successful application of the 
LSTM in a range of sequence prediction tasks, such as image captioning~\cite{DBLP:conf/cvpr/VinyalsTBE15} and 
machine translation~\cite{DBLP:conf/nips/SutskeverVL14}.

More formally, at the training phase, a training sample consists of an image $\Image$, annotated bounding boxes of a human body region (\eg, head region) of $N$ person instances in the image $\B = \{\b_1,\b_2,\ldots, \b_N\}$ ($N \geq 1$), and their corresponding labels $\Y = \{\y_1, \y_2, \ldots, \y_N\}$. Note that $N$ varies for different images. 

In our work, by treating both $\B$ and $\Y$ as sequences with some order
(the order for $\B$ and $\Y$ must be same, so an instance in $\B$ is
matched with its label in $\Y$), we aim to look for a set of parameters $\btheta^{\star}$ which maximizes the log likelihood of producing the correct label sequence 
$\Y$ given the input sequence $\B$ and the global image $\Image$ among all the training samples: 
\begin{equation}
\label{eq:theta}
\btheta^{\star} = \argmax_{\btheta} \sum_{(\B,\Image,\Y)} \log p(\Y|\B,\Image;\btheta).
\end{equation}
Intuitively, we can model the joint probability over all the labels
$\y_{1:N}$ using the chain factorization, that is, 
\begin{equation}
\label{eq:chain_rule}
\log p(\Y|\B,\Image;\btheta) = \sum_{t=1}^N \log p(\y_t|\y_{1:t-1},\b_{1:t},\Image;\btheta),  
\end{equation}
We assume that that the current prediction of $\y_t$  
does not depend on all instances in $\B$ but only the previous seen instances $\b_{1:t-1}$ and the current instance $\b_t$.

Analogous to sequence prediction models in other tasks~\cite{DBLP:conf/cvpr/VinyalsTBE15,DBLP:conf/nips/BengioVJS15}, we model the conditional probability  
$p(\y_t|\y_{1:t-1},\b_{1:t},\Image;\btheta)$ with a recurrent neural network,
by introducing a hidden state vector $\h_t$, that is, 
\begin{equation}
p(\y_t|\y_{1:t-1},\b_{1:t},\Image;\btheta) = p(\y_t | \h_t; \btheta).
\end{equation}
The hidden state vector $\h_t$ has the following form: 
\begin{equation}
\label{eq:recurrence}
\h_t = \left\{\begin{array}{ll}
  f(\Image;\btheta) & \mbox{if } t = 0, \\
  f(\h_{t-1}, \x_t;\btheta) & \mbox{otherwise.}
  \end{array}\right.
\end{equation}
The $\x_t$ in Eq.~\ref{eq:recurrence} is the novel part of our RNN  architecture, 
which is a joint embedding of previous label $\y_{t-1}$ and current 
input instance $\b_t$ (more details are provided in Sec.~\ref{subsec:joint_embed}). 
For $f()$ we opt for the LSTM component, which has shown state-of-the-art 
performance on sequence tasks such as image captioning~\cite{DBLP:conf/cvpr/VinyalsTBE15,
DBLP:conf/nips/BengioVJS15}.

As described in Eq.~\ref{eq:recurrence} and Fig.~\ref{fig:pipeline}, at the initial step ($t=0$), the input to the LSTM is the global 
image content $\Image$, which informs the network with scene context. For this purpose, we use features extracted from a Convolutional Neural Network (CNN) to represent images.  
In subsequent steps 
the inputs are the current joint embedding $\x_t$ and its previous hidden state $\h_{t-1}$ \footnote{When $t=1$, as there is no preceding identity labels, we add an auxiliary identity label $\y_0$.
This is similar to the case in image captioning~\cite{DBLP:conf/cvpr/VinyalsTBE15} where a special start token is used to represent the beginning of a sentence.}.

Let $\z_t$ denote the output of the LSTM at step $t$.
We then add a fully-connected layer ($\W$) 
with the softmax function on the top to generate $p(\y)$, the probability distribution over all the identity labels.

Our loss over all the  steps is the sum of the negative log likelihood of the ground-truth identity label $\y_t$ at each step:
\begin{equation}
L = -\sum_{t=1}^N \log p(\y_t). 
\end{equation} 
The above loss can be minimized by the Back-propagation Through Time (BPTT) technique. \\

\begin{figure}[t]
\centering
\begin{tabular}{@{}c@{}c}
\includegraphics[width=0.4\linewidth]{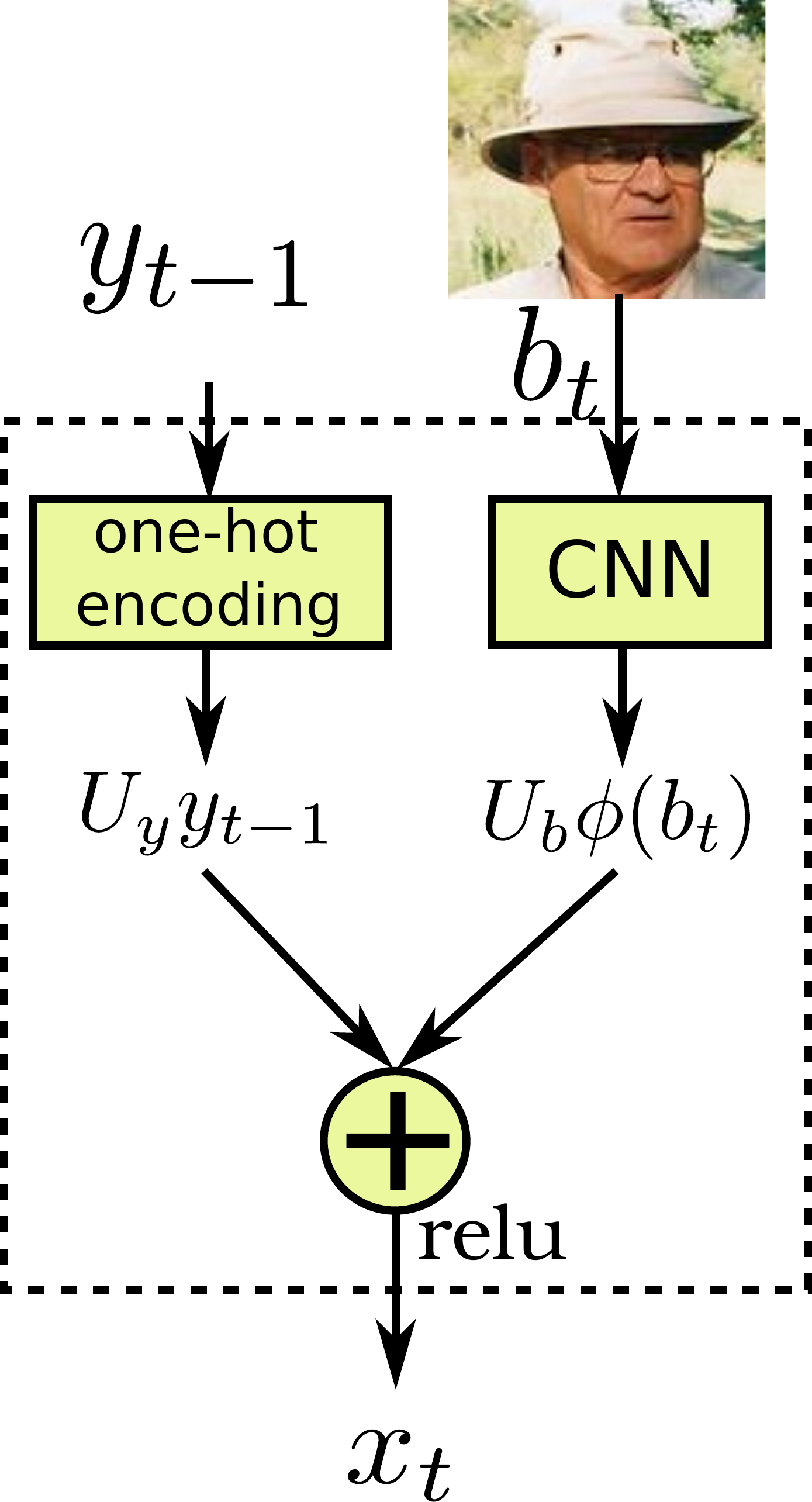} & \
\includegraphics[width=0.4\linewidth]{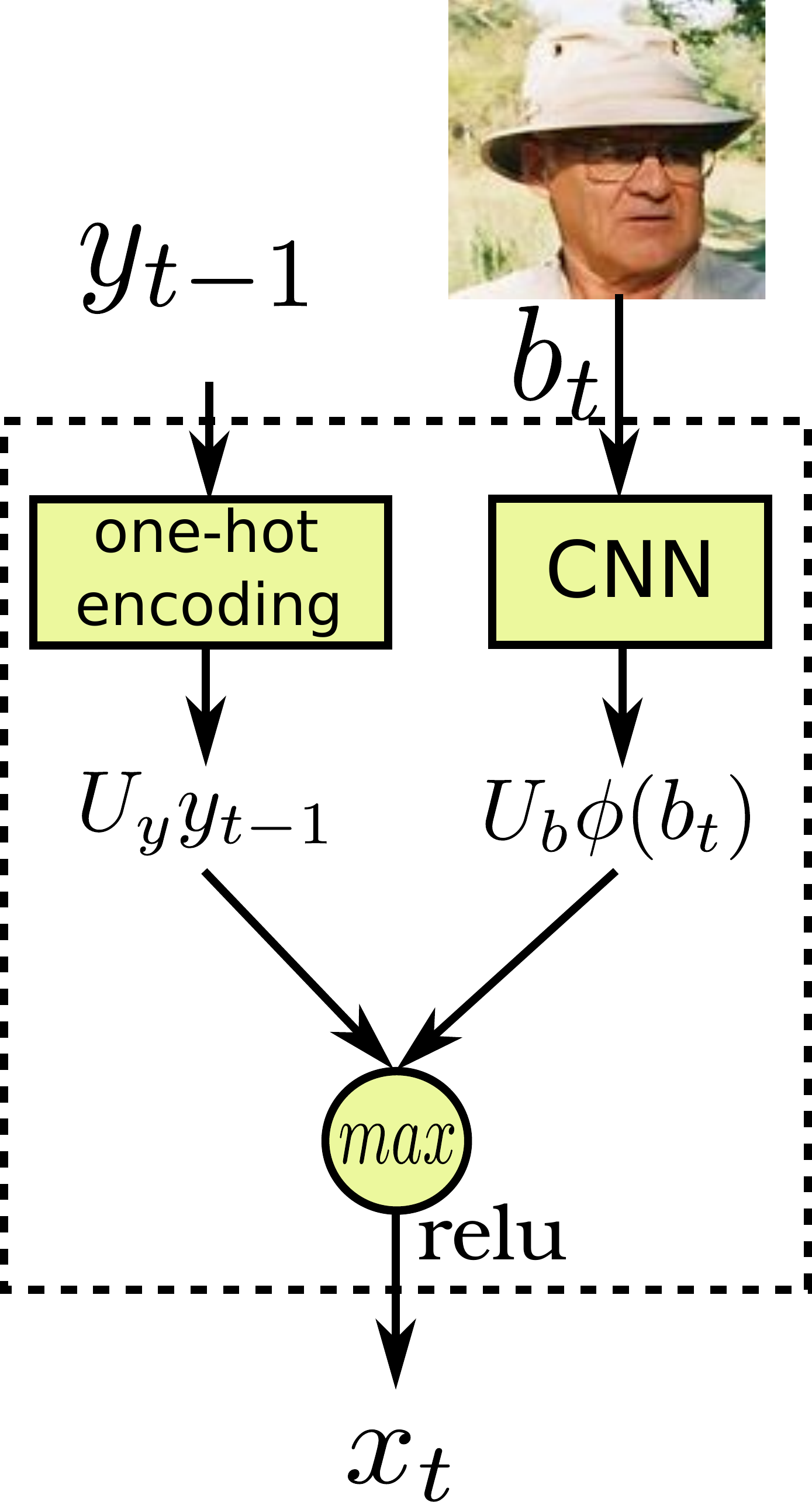} \ \\
(a) & (b)\\
\end{tabular}
\caption{Two variants of the joint embedding block in Fig.~\ref{fig:pipeline}. Either (a) addition or (b) an element-wise maximum can be used for joint embedding.}
\label{fig:joint_embed}
\end{figure}

\subsection{Joint embedding of instance feature and label}
\label{subsec:joint_embed} 
At any step $t$ ($t>0$), we argue that there are two sources of information which can help to predict the current label. 
The first source of information is the previous label $\y_{t-1}$ (exploiting label co-occurrence information). The other is the appearance of current instance $\b_t$, which is a feature vector denoted by $\bphi(\b_t)$. 
Therefore, we propose to introduce a joint embedding $\x_t$ to combine the information from the two sources. 
Specifically, as shown in Fig.~\ref{fig:joint_embed}, after transforming labels into one-hot vectors, we define two embedding matrices $\U_y$ and $\U_b$ for encoding $\y_{t-1}$ and $\bphi(\b_t)$ respectively:
\begin{equation}
\label{eq:addition}
\x_t = \relu(\U_y \y_{t-1} + \U_b \bphi(\b_t)), 
\end{equation}
where $relu$ stands for the Rectified Linear (ReLU) activation function. 
This is motivated by the label embedding formulation of~\cite{DBLP:conf/cvpr/WangYMHHX16}.

An alternative to the addition above is to take the element-wise maximum with the ReLu activation, that is, 
\begin{equation}
\label{eq:max}
\x_t = \relu( \max (\U_y \y_{t-1}, \U_b \bphi(\b_t)).
\end{equation}
The performance of these two formulations are analyzed in the experiment section (Sec.~\ref{subsec:ablation}).\\

\subsection{Training and inference}
\label{subsec:training_inference}

\noindent\textbf{Random order training.}
In some sequence prediction models in Computer Vision, such as 
image captioning, there is a natural order of the input sequence 
(\eg, words in a sentence). 
For some tasks when the order is not obvious, the order is pre-defined 
based on some heuristic rules. 
For instance, in the human pose estimation work of~\cite{DBLP:conf/eccv/GkioxariTJ16}, a tree based ordering of joints is used. 
The investigation of Vinyals~\etal~\cite{DBLP:journals/corr/VinyalsBK16} 
has shown that for some simple problems most 
orderings perform equally well.
Our task differs from the 
sequence prediction tasks above, as people appearing in images don't have an inherent order.  
We thus select a random order at training time.
To be more specific, for a training image, its annotated instances and their identity labels are randomly shuffled in the same order to generate a input sequence and a target sequence respectively. 
Therefore, the order of people in a training image varies in different epoches, which incorporates randomness to the training process. \\

\begin{figure}[t]
\begin{center}
\includegraphics[width=0.9\linewidth]{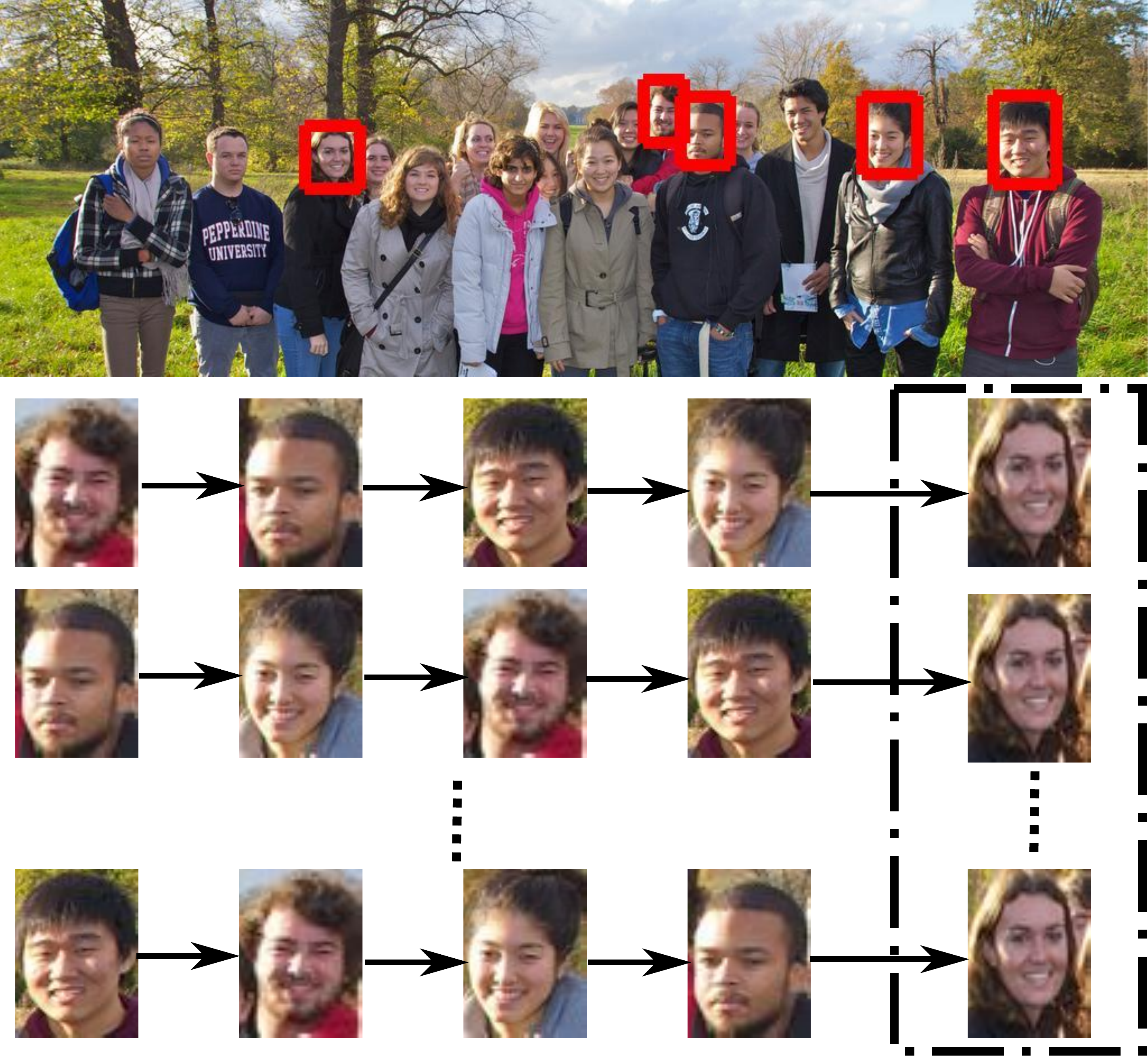} \\
\end{center}
\caption{At test time, to predict the label of a query instance (\eg, the one in the dashed box), we 
generate multiple sequences for this instance, all with the instance at the end and other instances randomly ordered. The probability of an identity label for this instance is the maximum probability of this label in all sequences. 
}
\label{fig:inference} %
\end{figure}

\noindent\textbf{Inference.} At test time, to predict the identity label of a query instance, we generate multiple sequences for this instance,  
all with the query instance at the end and other instances in the image randomly ordered.
The rationale behind this is that in order to take the advantage of the rich relational information between all people appearing in the same image, our sequence prediction model 
should ``see'' all other instances in the image before predicting the query instance. 
Fig.~\ref{fig:inference} provides a demonstration of the inference process.  
Note that the same process is done for every instance in a test image.

For each of the sequences for a query instance, 
we first feed the global image content to get initial state of the LSTM. 
The labels at subsequent steps are then predicted deterministicaly. 
More specifically, at step $t$ ($t>0$), the predicted label is the identity with the maximum output probability, \ie, $\y_t = \argmax p(\y)$, which is 
then used as the input label for joint embedding with another randomly selected instance at the next step. 
This process stops after the query instance has been processed, which results in a probability distribution over identities for the query instance at the end of the sequence. 

Given all the probability distributions from different sequences of the query instance, we take the element-wise maximum of the probability distributions and then the identity label with the maximum probability after this operation is assigned to the query instance. \\

\subsection{Discussion}
\label{subsec:disscussion}

\noindent\textbf{Context information.} Contextual cues beyond 
human appearance have been 
found valuable for person recognition in photo albums~\cite{DBLP:conf/iccv/OhBFS15,DBLP:conf/cvpr/LiBLSH16}.
There are two types of contextual cues exploited 
in our framework. 

The first type of contextual cue is the relation context, which is the relational information between people in the same image, \ie, some people tend to appear together in the same 
image. 
In our work, this relational information is naturally 
handled by our sequence prediction formulation using an LSTM. 
Note that the relational information captured in 
our work is not the simple label co-occurrence between people, as our LSTM receives a joint embedding of both label and appearance as input.

The second type of contextual cue is the scene context. 
This is based on the assumption that some identities appear more frequently in some certain scenes. 
Thus the scene context can be utilized as a prior for indicating which identities are likely to appear. 
In our work, this is done by feeding the global image feature to the LSTM at the initial step. 

As we will show later in the experiments (Sec.~\ref{subsec:ablation}), the above contextual cues are vital for improving the classification accuracy.\\

\noindent\textbf{Multiple body regions.} So far we only assume that 
$\B$ is from a single annotated body part (\eg, head). 
Recent work on person recognition in photo albums~\cite{DBLP:conf/cvpr/ZhangPTFB15,DBLP:conf/iccv/OhBFS15,
DBLP:conf/cvpr/LiBLSH16} has shown that recognition performance 
can be improved by fusing information from multiple body regions. 
In this work, we also extend our formulation to the multiple region case. 
One obvious solution is  feature concatenation across multiple regions. 
We provide in-depth analysis of some feature fusion methods in the experiments section (Sec.~\ref{subsec:region}). \\

%% file: experiment.tex
\begin{table}[t]
\begin{center}
\scalebox{0.9}{
\begin{tabular}{|l|c|c|c|c|c|}
\hline
\multirow{2}{*}{Test split} & \multirow{2}{*}{$\#$identities} & \multicolumn{2}{|c|}{$\#$instances} &
 \multicolumn{2}{|c|}{$\#$multi instances}\\
\cline{3-6}
 & & $test_0$ & $test_1$ & $test_0$ & $test_1$\\
\hline
\hline
Original & $581$ & $6442$ & $6443$ & $2802$ & $2797$\\
\hline
Album & $581$ & $6497$ & $6388$ & $2814$ & $2751$ \\
\hline
Time & $581$ & $6440$  & $6445$ & $2591$ & $2647$\\
\hline
Day  & $199$ & $2484$ & $2485$ & $792$  & $744$\\
\hline
\end{tabular}
}
\end{center}
\caption{Statistics of the four test splits of the PIPA dataset. 
``$\#$identities'' denotes the number of identity labels,  
``$\#$instances'' refers to the number of all instances, and `$\#$multi instances'' is the number of instances from images contain 
multiple instances.}
\label{tab:pipa}
\end{table}

\subsection{Experimental setup}
\noindent\textbf{Dataset and evaluation metric.} The \emph{People In Photo Albums (PIPA)} dataset~\cite{DBLP:conf/cvpr/ZhangPTFB15} is adopted for evaluation of our approach as well as some baselines. 
The PIPA dataset is partitioned intro train, validation, test, and leftover sets. 
The head region of each instance has been annotated in all sets (see Fig.~\ref{fig:example}).
In previous work~\cite{DBLP:conf/cvpr/ZhangPTFB15,DBLP:conf/iccv/OhBFS15,
DBLP:conf/cvpr/LiBLSH16}, the training set is used only for learning 
good feature representations for body regions. 
In the standard evaluation setting proposed in
~\cite{DBLP:conf/cvpr/ZhangPTFB15}, the test set itself is split into 
two subsets, $test_0$ and $test_1$ with roughly the same number of 
instances. 
Given a recognition system that is trained on $test_0$, it is then evaluated on $test_1$, and vice visa. 

\begin{table*}[t]
\begin{center}
\begin{tabular}{|l|c|c c|c|c|c|}
\hline
Split & Method & Relation 
& Scene & Acc overall (\%) & Acc multi (\%) & Acc single (\%)\\
\hline 
\multirow{3}{*}{Original}
& Appearance-only & -- & -- & $75.43$ & $77.93$ & $73.51$\\
\cline{2-7}
& Ours-relation & \checkmark  & -- & $76.73$ & $80.73$ & $73.66$\\
\cline{2-7}
& Ours & \checkmark & \checkmark & \boldsymbol{$81.75$} & \boldsymbol{$84.85$} & \boldsymbol{$79.36$}\\
\hline
\hline
\multirow{3}{*}{Album}
& Appearance-only & -- & -- & $68.31$ & $72.00$ & $ 65.52$\\
\cline{2-7}
& Ours-relation & \checkmark  & -- & $68.85$ & $73.41$ & $65.38$\\
\cline{2-7}
& Ours & \checkmark & \checkmark & \boldsymbol{$74.21$} & \boldsymbol{$78.22$} & \boldsymbol{$71.16$}\\
\hline
\hline
\multirow{3}{*}{Time}
& Appearance-only & -- & -- & $57.19$  & $59.79$ & $55.39$\\
\cline{2-7}
& Ours-relation & \checkmark  & -- & $58.57$ & $63.19$ & $55.39$\\
\cline{2-7}
& Ours & \checkmark & \checkmark & \boldsymbol{$63.73$} & \boldsymbol{$67.17$} & \boldsymbol{$61.37$}\\
\hline
\hline
\multirow{3}{*}{Day}
& Appearance-only & -- & -- & $36.37$ & $36.72$ & $36.23$\\
\cline{2-7}
& Ours-relation & \checkmark  & -- & $40.39$ & $ 44.25$ & $38.71$\\
\cline{2-7}
& Ours & \checkmark & \checkmark & \boldsymbol{$42.75$} & \boldsymbol{$47.25$} & \boldsymbol{$40.74$}\\
\hline
\end{tabular}
\end{center}
\caption{Classification accuracy($\%$) of two baselines as well as our full system under four settings on the PIPA test set.``Acc overall'' refers to classification accuracy computed from all instances, whereas ``Acc multi'' (\emph{resp.} ``Acc single'') refers to accuracy on instances from images contain multiple (\emph{resp.} single) instances. Head region is adopted for this analysis. 
Clearly, by modeling both relation and scene context, our full system 
has outperformed two baselines by a noticeable margin.   
}
\label{tab:ablation}
\end{table*}

Recently, in addition to the original test split proposed in
~\cite{DBLP:conf/cvpr/ZhangPTFB15}, three more challenging splits are introduced in~\cite{DBLP:conf/iccv/OhBFS15}, including 
album, time and day splits. 
The album split ensures instances in $test_0$ and $test_1$ are from different albums, while time and day spilts emphasis the temporal distance between $test_0$ and $test_1$ (different events, different days \etc). Generally speaking, the ranking of these four spilts in the order of increasing difficulty is: original, album, time and day 
split.
We provide an overview of the statistics of the four test splits in 
Table~\ref{tab:pipa}.

Following previous work~\cite{DBLP:conf/cvpr/ZhangPTFB15,DBLP:conf/iccv/OhBFS15,
DBLP:conf/cvpr/LiBLSH16}, classification accuracy is used to evaluate the performance our approach and some baselines. 
Specifically, the average classification accuracy over the $test_0$ and $test_1$ is reported. 
To have a in-depth understanding of our system, we also report 
the average accuracy on instances from images with multiple and single instances respectively.\\

\noindent\textbf{Implementation details.} 
Two body regions, head and upper body, are exploited
in our system. 
Based on bounding box annotations of the head region which have been provided by the PIPA dataset, we estimate the annotations for the  upper body region, similar to~\cite{DBLP:conf/iccv/OhBFS15}. 
To learn feature representations, we 
fine-tune two VGG-16 networks~\cite{DBLP:journals/corr/SimonyanZ14a} on the PIPA training set for these two regions respectively. 
On the test splits ($test_0$ and $test_1$), we extract CNN features from the $fc7$ layer of the fine-tuned networks for the two regions 
respectively.
The global image content $I$ is the 4096-D $fc7$ feature extracted from the vanilla VGG-16 network, which is pre-trained on the 
ImageNet~\cite{DBLP:conf/cvpr/DengDSLL009}.  

As each image may contain different numbers of instances, we unroll the LSTM to a fixed 22 steps (the maximum 
 number of instances can appear on the PIPA test splits) during training. 
For images with instances less than 22, we pad labels with zeros and do not calculate loss on the padded labels.
 
We train all our weights, including the label embedding weight $\U_y$, the image embedding weight $\U_b$, the classification weights $\W$, and weights in the LSTM, and using stochastic gradient descent with the Adam optimizer~\cite{DBLP:journals/corr/KingmaB14}. 
The initial learning rate is setted as 0.001 and decreased by 10 times after 20 epochs. We stop training after 80 epochs. 
We use 512 dimensions for the embeddings and the size of the LSTM memory.

\subsection{Ablation study}
\label{subsec:ablation}

To investigate the impact of the different components of our approach, 
we consider the following two baselines as well as our full system.
\begin{enumerate}
\itemsep -1mm
\item ``Appearance-only'': given the CNN network fine-tuned on a body region on the PIPA training set, we fine-tune the last fully-connected layer on the test instances on the either of the two test splits and evaluate on the other. In this sense, this setting is only based on the visual appearance of identities without any contextual information.
\item ``Ours-relation'': We still use our LSTM-based framework to model
the relation context between people, but we do not feed the LSTM with in global image content $I$ at the initial step. 
In other words, the scene context is not exploited in this setting. 
Therefore, the only contextual cue exploited in this setting is the 
relation context. 
    
\item ``Ours'': Our full sequence prediction model with the global image content $I$ fed at the initial step. 
In this sense, both relation and scene context are exploited. 

\end{enumerate} 

Table.~\ref{tab:ablation} depicts the performance of these three approaches under four different settings.\\

\noindent\textbf{The importance of relation context.}
Comparing the overall performance of ``Our-relation'' with that of the ``Appearance-only'' (Table.~\ref{tab:ablation}), we observe that the former bypasses the later in all of the four settings. 
This reflects that our sequence prediction model has successfully taken the advantage of relation information between multiple people in the same photo.

When taking a closer look at the result of the two baselines in terms of the ``Acc multi'', it is clear that the ``Acc multi'' has substantial increases in the ``Our-relation'' case, which contributes to the improvements in overall accuracy.
We also observe that ``Acc single'' has stayed stable between the ``Appearance-only'' and ``Our-relation'' cases.  
This is understandable because when there is only one instance in the image, there is no relation context to be exploited.\\

\noindent\textbf{The importance of scene context.} 
Comparing the performance of our full system with that of the ``Our-relation'' case (Table.~\ref{tab:ablation}), 
we observe that feeding the global image content to the our sequence prediction model always
leads to further substantial improvements, resulting in 
our best performance in all the four settings. 

More specifically, in all of the four settings, in terms of the overall accuracy, there is about $3\sim4\%$ accuracy gain by using our full system, compared with 
that of the  ``Our-relation'' baseline which does not take advantage of the scene context. 
Similar improvements are also observed in the both ``Acc multi'' and 
``Acc single''. 
This verifies that the scene context is valuable for person recognition.

Also the recent work of~\cite{DBLP:conf/iccv/OhBFS15} has found that scene contains useful information for person recognition, thus the observation in our work is consistent with~\cite{DBLP:conf/iccv/OhBFS15}. 
In contrast with~\cite{DBLP:conf/iccv/OhBFS15} in which the global image cue is analyzed independent of other cues, 
we are the first to incorporate different contextual cues, along with instances' visual features into an end-to-end trainable framework. \\

\noindent\textbf{Joint embedding analysis.} As depicted in Fig.~\ref{fig:joint_embed}, we have proposed two formulations for the
joint embedding of instance feature and label, in which 
information from the two sources is fused either by 
addition (Eq.~\ref{eq:addition}) or element-wise max (Eq.~\ref{eq:max}).
We hereby analysis the performance of these two variants on the PIPA test set. 

\begin{table}[t]
\begin{center}
\begin{tabular}{|l|c|c|}
\hline
Test split & Addition & Element-wise max\\
\hline
\hline 
Original & $81.51$ & \boldsymbol{$81.75$}\\
\hline 
Album & $73.21$ & \boldsymbol{$74.21$}\\
\hline
Time & $62.97$ & \boldsymbol{$63.73$}\\
\hline 
Day & \boldsymbol{$43.15$} & $42.75$\\
\hline
\end{tabular}
\end{center}
\caption{Classification accuracy ($\%$) of the two variants of joint embedding formulation in Fig.~\ref{fig:joint_embed}. Head region is used in this case.}
\label{tab:joint_embed}
\end{table}

As depicted in Table.~\ref{tab:joint_embed}, both embedding formulation shows very close performance in the four settings, 
with the ``Max'' fusion slightly bypassing the ``Addition'' fusion in the three out of four settings. 
Therefore, we opt for the``Max'' fusion method to report results in the following.

\begin{table}[t]
\begin{center}
\begin{tabular}{|l|c|c|c|c|c|}
\hline
\multirow{2}{*}{Test split} & \multicolumn{2}{|c|}{Single region}
& \multicolumn{3}{|c|}{Multiple region fusion}\\
\cline{2-6}
& Head & Upper & Avg & Max & Concat\\
\hline 
\hline
Original & \boldsymbol{$81.75$} & $79.92$ & \boldsymbol{$84.93$} & $84.07$ & $82.86$\\
\hline 
Album & \boldsymbol{$74.21$} & $70.78$ & \boldsymbol{$78.25$} & $75.88$ & $74.66$\\
\hline
Time & \boldsymbol{$63.73$} & $58.80$ & \boldsymbol{$66.43$} & $65.63$ & $63.62$\\
\hline 
Day & \boldsymbol{$42.75$} & $34.61$ & \boldsymbol{$43.73$} & $43.55$ & $41.56$\\
\hline
\end{tabular}
\end{center}
\caption{Classification accuracy ($\%$) using a single body region (columns 2-3), and their different fusions (columns 3-5).}
\label{tab:region_analysis}
\end{table}

\begin{table*}[t]
\begin{center}
\begin{tabular}{|l|c|c|c|c|c|c|}
\hline
 \multirow{2}{*}{Test split} & \multicolumn{2}{|c|}{Head} &
 \multicolumn{2}{|c|}{Upper body} &
 \multicolumn{2}{|c|}{Head + Upper body}\\
\cline{2-7}
& \cite{DBLP:conf/iccv/OhBFS15} & Our & \cite{DBLP:conf/iccv/OhBFS15} & Our & \cite{DBLP:conf/iccv/OhBFS15} & Our\\
\hline
\hline
Original & $76.42$ & \boldsymbol{$81.75$} & $75.07$ & \boldsymbol{$79.92$} & $83.63$ & \boldsymbol{$84.93$}\\
\hline
Album & $67.48$ & \boldsymbol{$74.21$} & $64.65$ & \boldsymbol{$70.78$} & $74.52$ & \boldsymbol{$78.25$}\\
\hline
Time & $57.05$  & \boldsymbol{$63.73$} & $50.90$ & \boldsymbol{$58.80$} & $63.98$ & \boldsymbol{$66.43$}\\
\hline
Day & $36.37$ & \boldsymbol{$42.75$} & $24.17$ & \boldsymbol{$34.61$} & $38.94$ & \boldsymbol{$43.73$}\\
\hline
\end{tabular}
\end{center}
\caption{Classification accuracy($\%$) of Oh~\etal~\cite{DBLP:conf/iccv/OhBFS15} and ours under the four different settings on the PIPA test set, using head region, upper body region and their fusion.  
}
\label{tab:state-of-the-arts}
\end{table*}

\subsection{Region analysis}
\label{subsec:region}

Recent studies on person recognition in photo albums~\cite{DBLP:conf/cvpr/ZhangPTFB15,DBLP:conf/iccv/OhBFS15,
DBLP:conf/cvpr/LiBLSH16} have shown different human body 
regions are complementary for achieving good performance. 
Thus, a ensemble of models built on different body regions 
are adopted in previous works.
For instance, 107 poselet~\cite{DBLP:conf/iccv/BourdevM09} classifiers
are used by Zhang~\etal~\cite{DBLP:conf/cvpr/ZhangPTFB15}.
Oh~\etal~\cite{DBLP:conf/iccv/OhBFS15} have analyzed the contribution 
of different body regions (\eg, face, head, upper body, full body) to the recognition performance. 
Following the above works, we also extend our model to handle multiple body regions. 
In particular, the two body regions exploited in our work are the head and 
upper body regions. \\

\noindent\textbf{Head vs. Upper body.} %
As shown in Table.~\ref{tab:region_analysis}, 
the usage of the head region outperforms the upper body region 
in all the settings, and the gap increases when the dataset comes 
more challenging (\eg, time or day setting). 
This is understandable because in the time or day setting, the appearance of the upper body of same person can have significant changes (\eg, change of clothing), which results in more failures when 
using features from the upper body region. 
In comparison, features from the head region are relative stable. \\

\noindent\textbf{Multiple region fusion.} 
As different body regions have different levels of relative importance 
for the final recognition performance, the information from different regions should be fused in order to achieve better performance. 
In our work, we study three methods to fuse information from head and upper body regions in the our sequence prediction model.

\begin{enumerate}
\itemsep -1mm 
\item ``Concat''. This is our model trained with concatenated features from head and upper body regions. 
\item ``Avg''. Two models are trained with features from 
the head and upper body region respectively. 
The probability of a test instance is the average of probabilities 
from the two models.

\item ``Max''. Same as the ``Avg'' case except that we take the maximum of probabilities rather than average.

\end{enumerate}

The performance of the above three methods are presented in Table.~\ref{tab:region_analysis}. 
Clearly, the ``Avg'' fusion method shows the largest improvement 
over the performance of using a single head or upper body region. 
Therefore, we use the performance of ``Avg'' fusion method to compare with state-of-the-art approaches in the following.

\begin{figure*}[t]
\vspace{-0.0cm}
\begin{center}
\begin{tabular}{@{}c@{}c@{}c@{}c@{}c}
\includegraphics[width=0.2\linewidth, height=2.3cm]{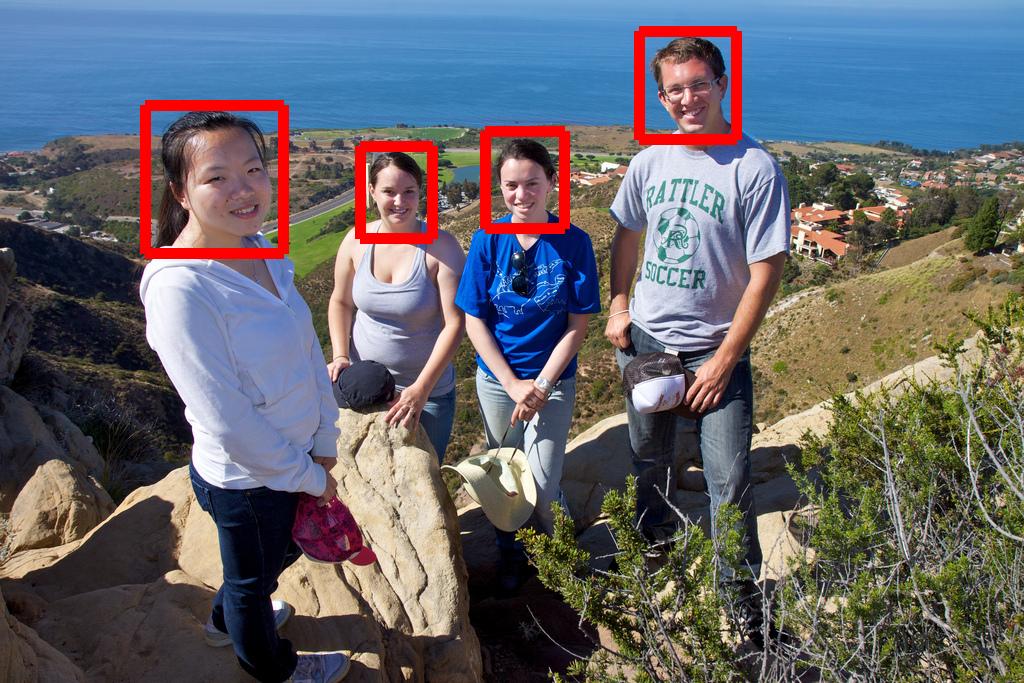} \ &
\includegraphics[width=0.2\linewidth, height=2.3cm]
{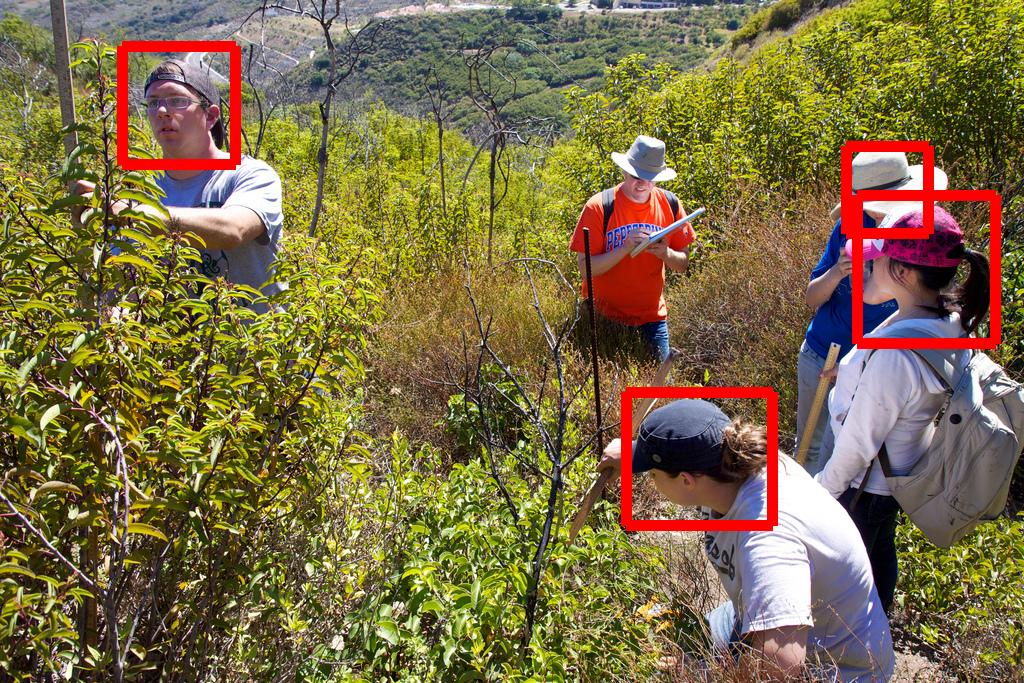} \ &
\includegraphics[width=0.2\linewidth, height=2.3cm]{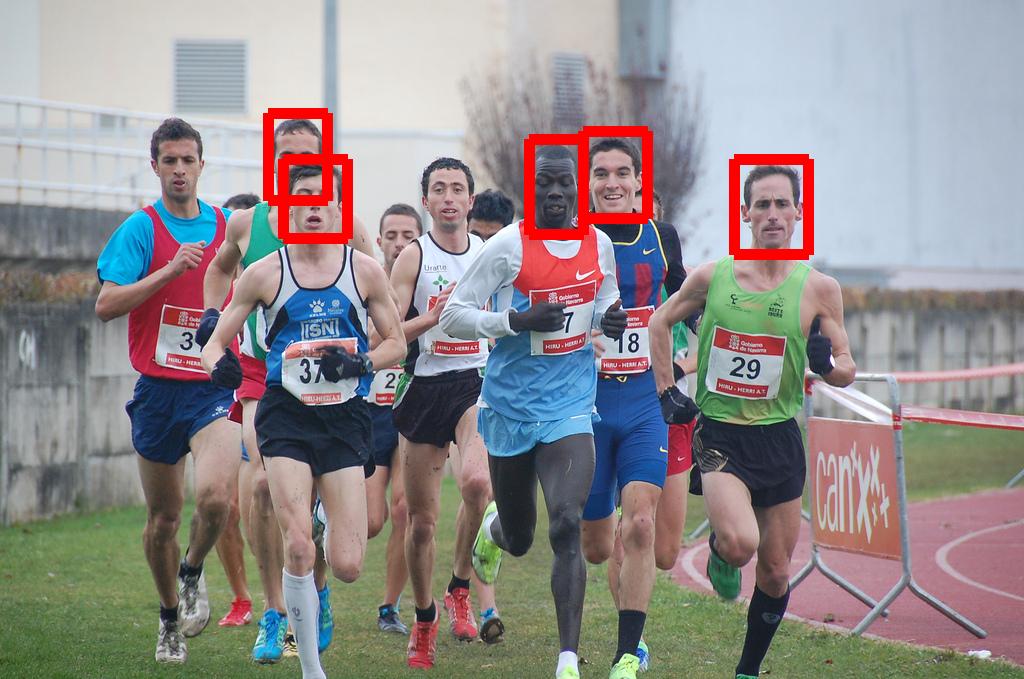} \ &
\includegraphics[width=0.2\linewidth, height=2.3cm]{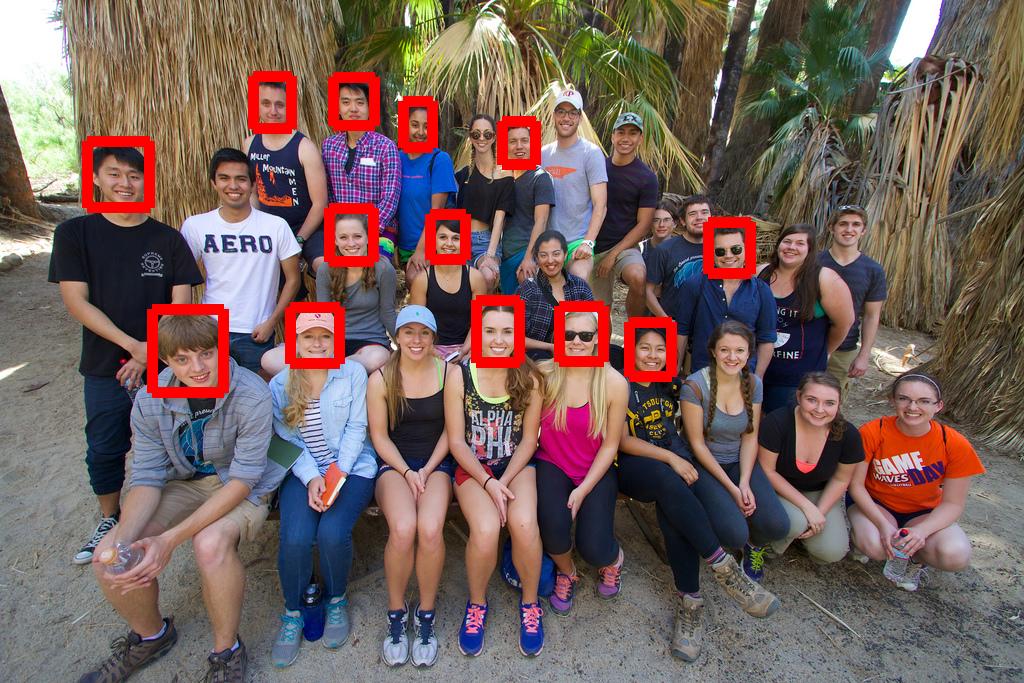} \ &
\includegraphics[width=0.2\linewidth, height=2.3cm]{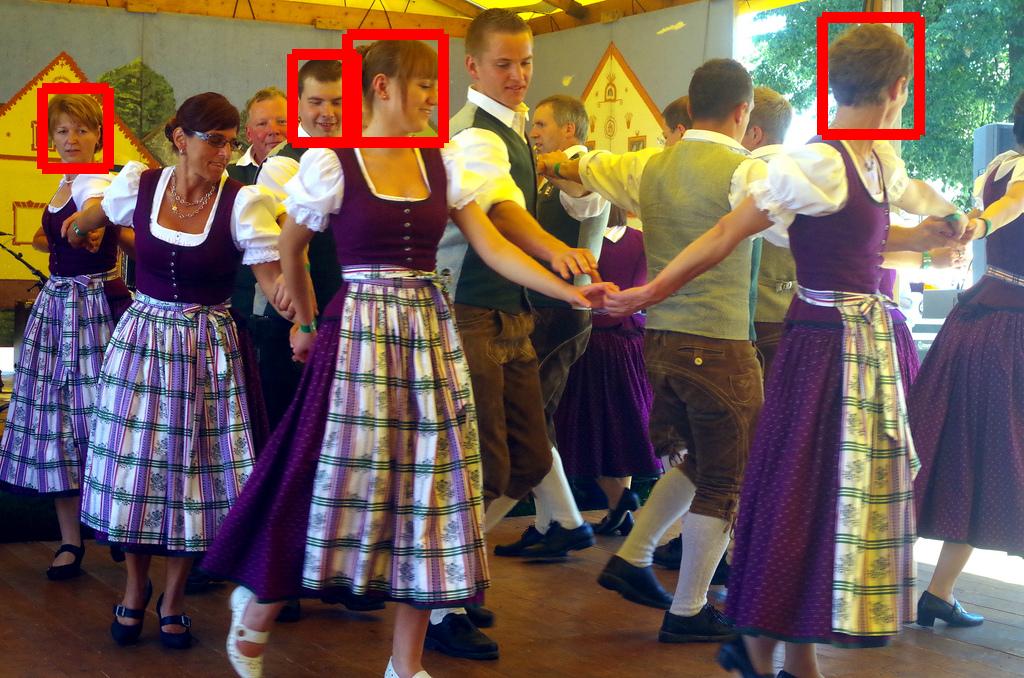} \ \\
\includegraphics[width=0.2\linewidth, height=2.3cm]{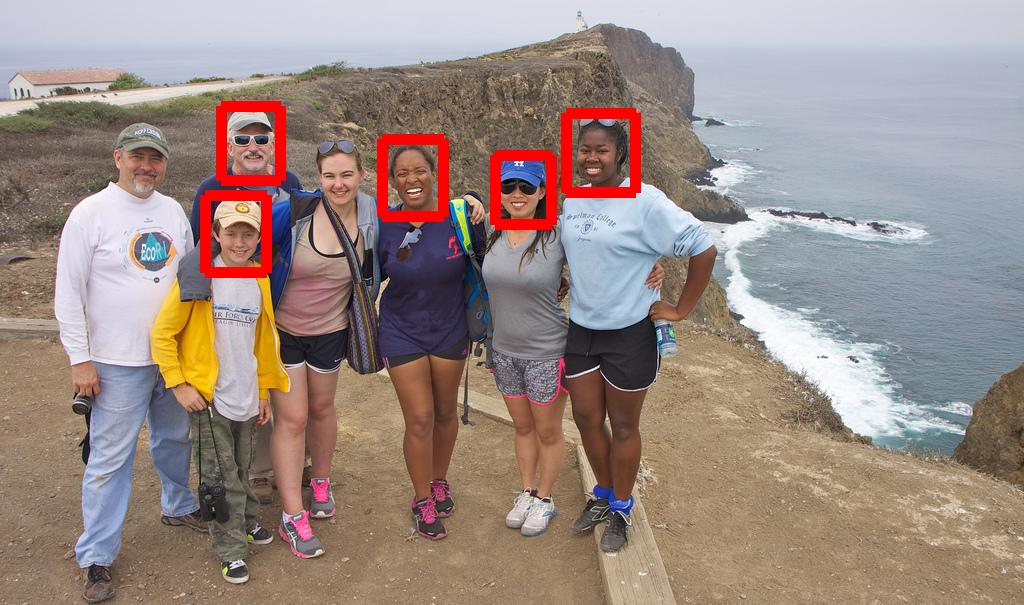} \ &
\includegraphics[width=0.2\linewidth, height=2.3cm]{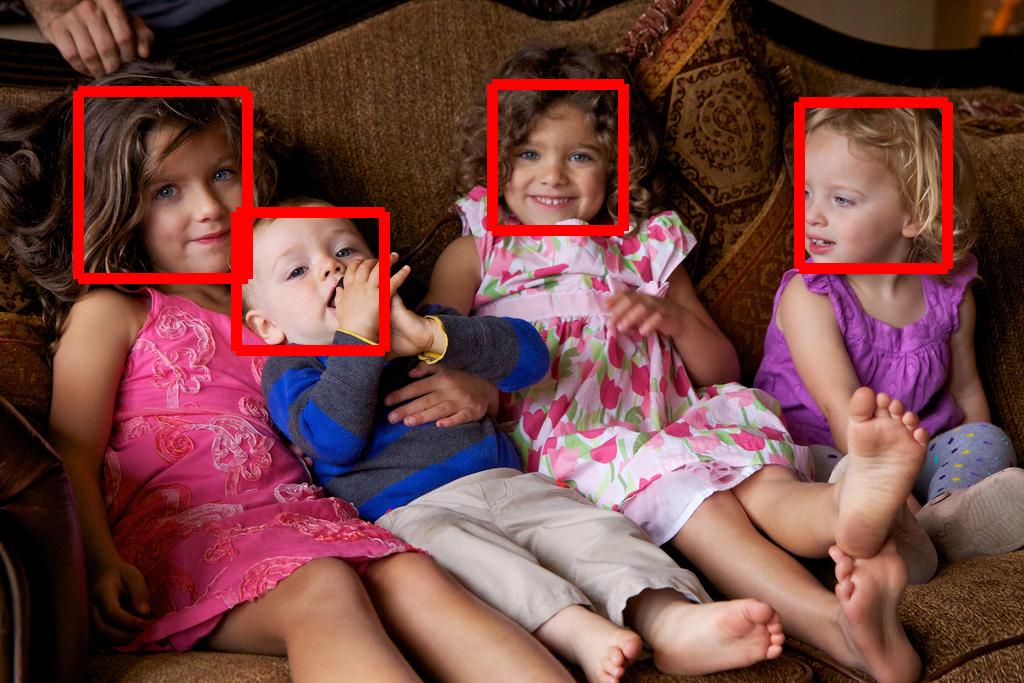} \ &
\includegraphics[width=0.2\linewidth, height=2.3cm]{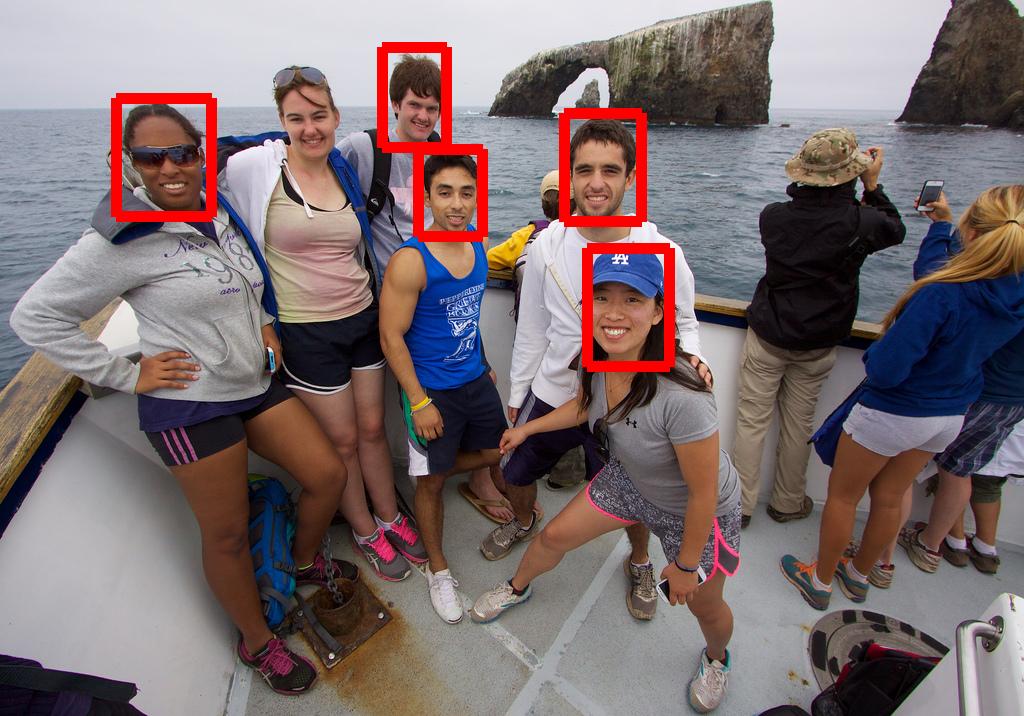} \ &
\includegraphics[width=0.2\linewidth, height=2.3cm]{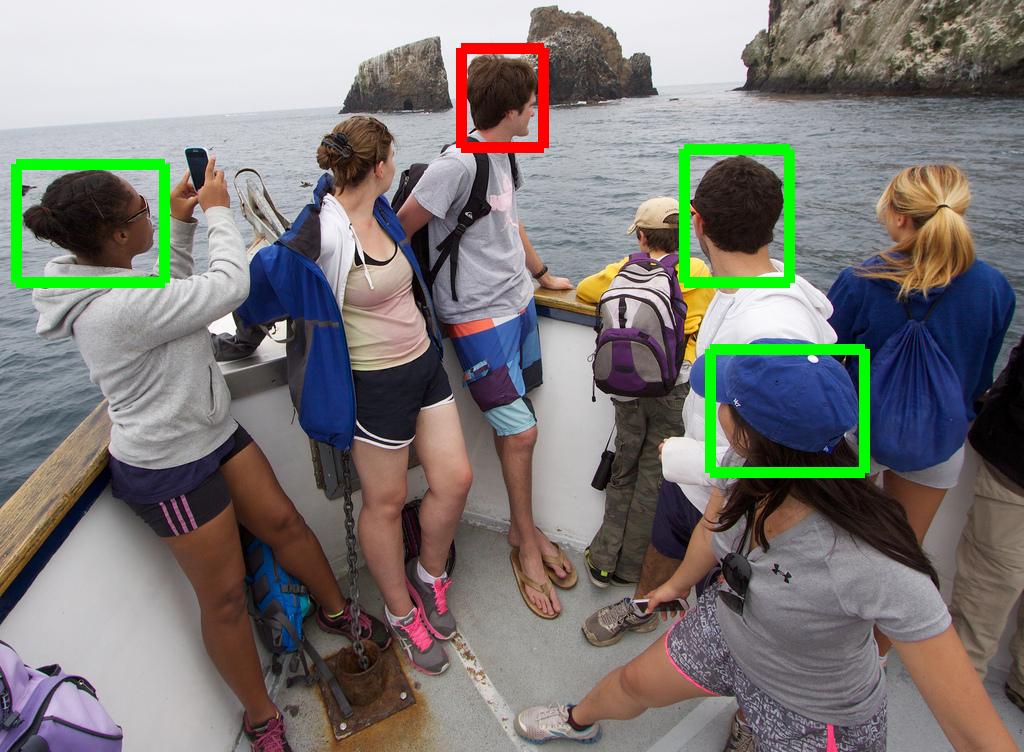} \ &
\includegraphics[width=0.2\linewidth, height=2.3cm]{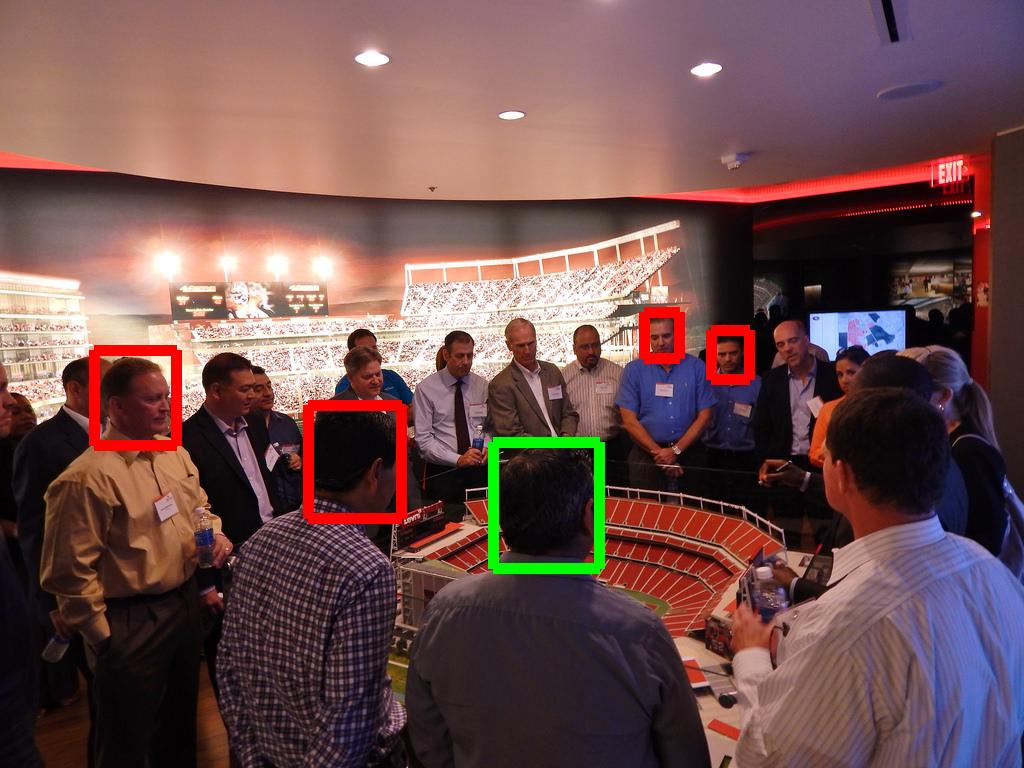} \ \\
\includegraphics[width=0.2\linewidth, height=2.3cm]{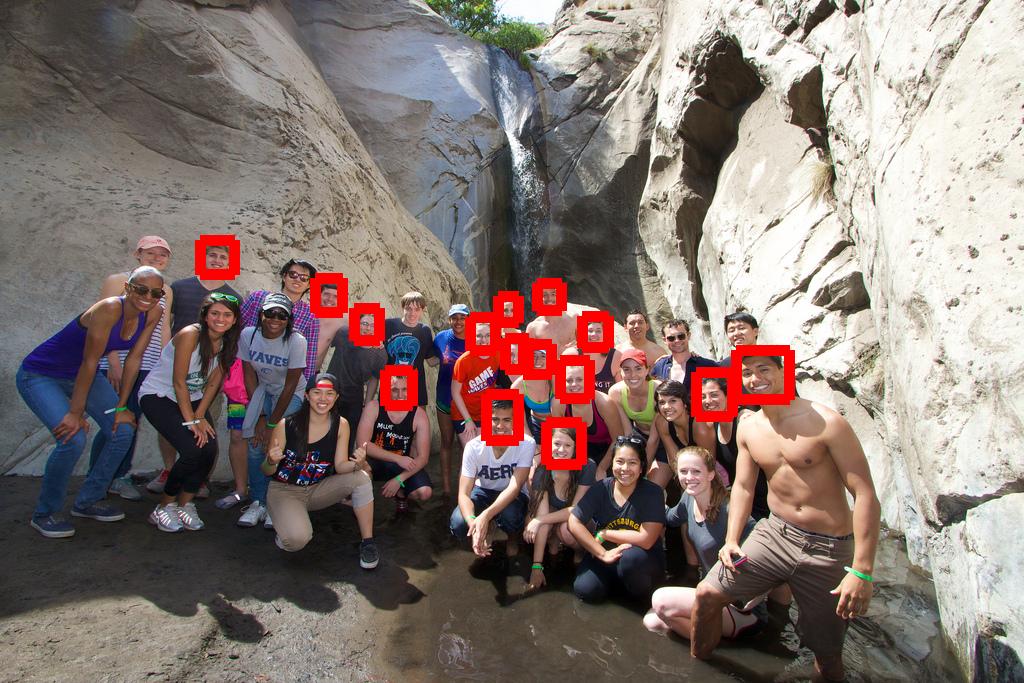} \ &
\includegraphics[width=0.2\linewidth, height=2.3cm]{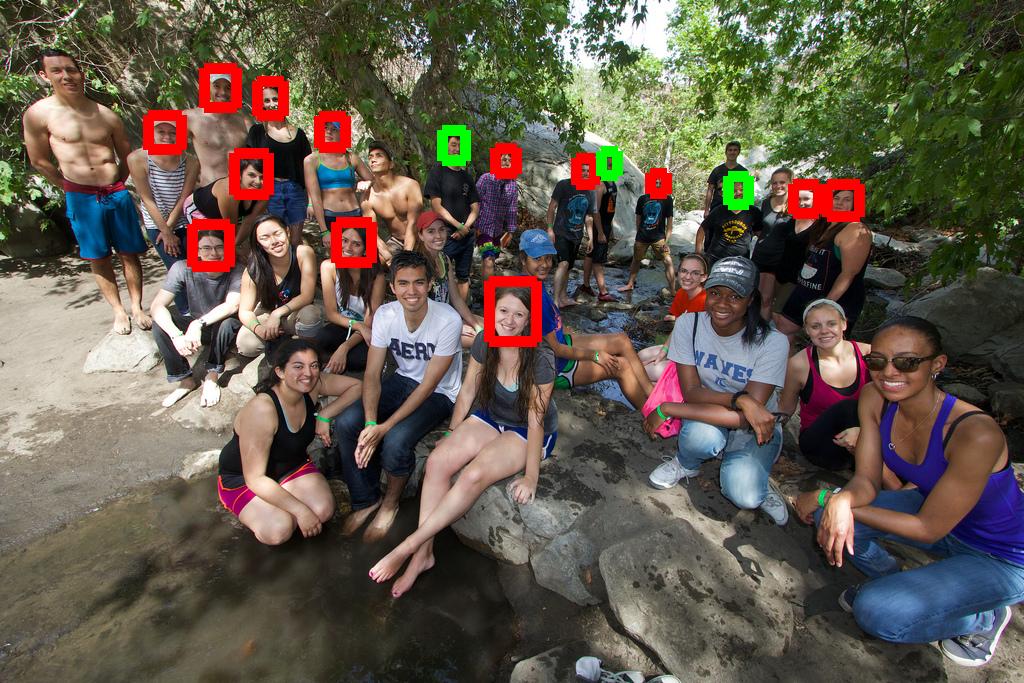} \ &
\includegraphics[width=0.2\linewidth, height=2.3cm]{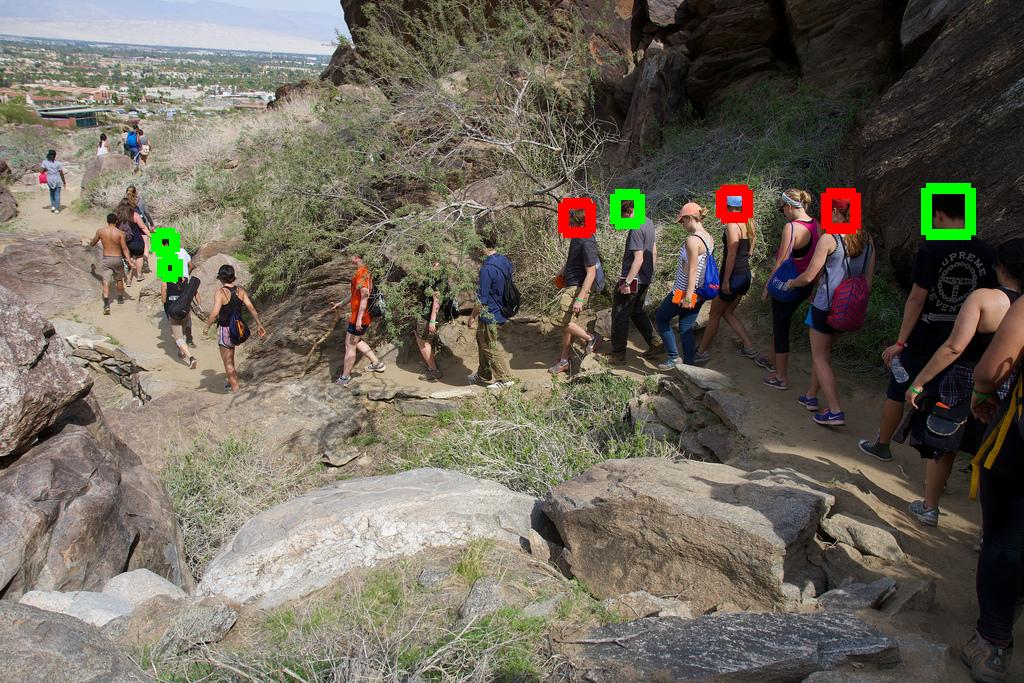} \ &
\includegraphics[width=0.2\linewidth, height=2.3cm]{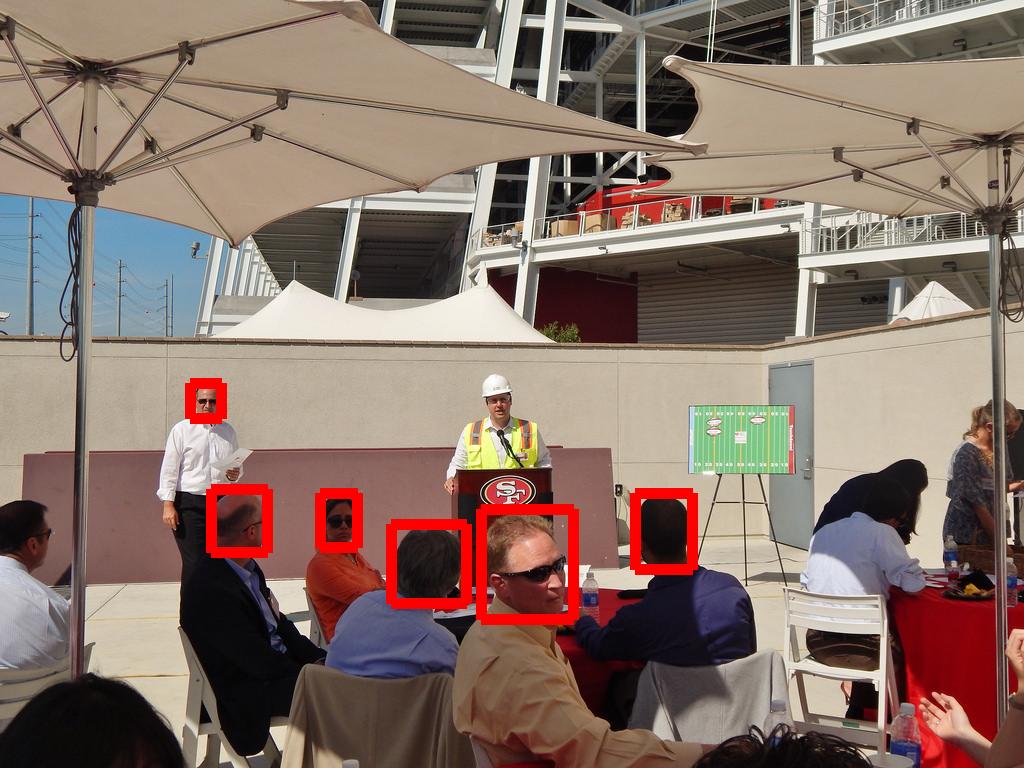} \ &
\includegraphics[width=0.2\linewidth, height=2.3cm]{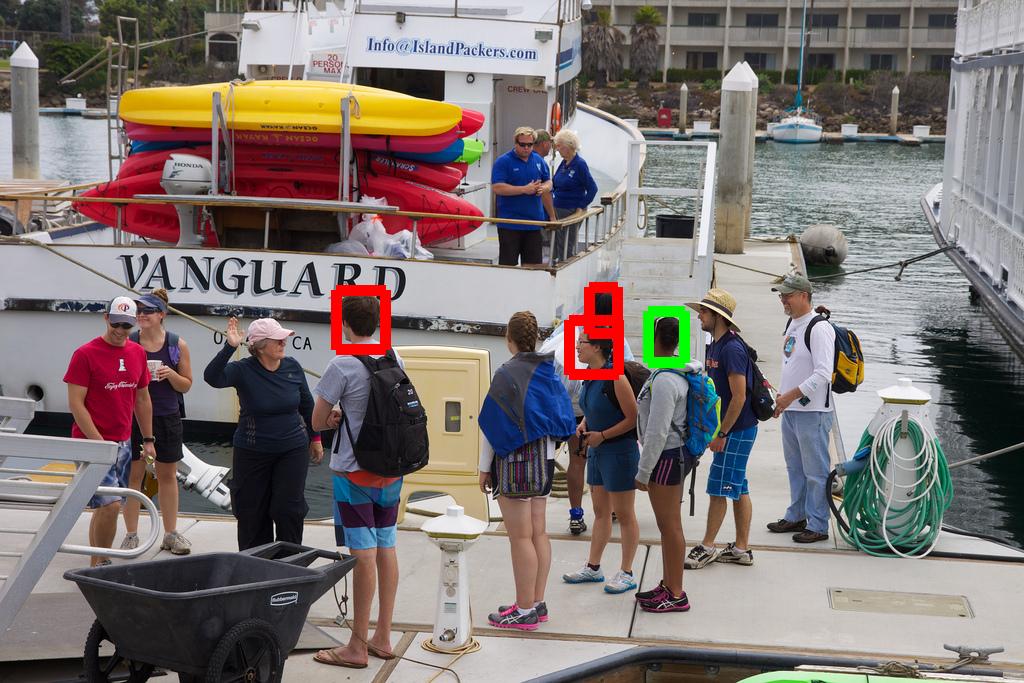} \ \\
\end{tabular}
\end{center}
\caption{Our predictions using the head region on the PIPA test set. Correct predictions are denoted by red bounding boxes whereas wrong ones are in green. Best viewed in color.}
\label{fig:examples}
\end{figure*}

\subsection{Comparison to state-of-the-arts}
We now compare the performance of our approach to the state-of-the-art
approaches in person recognition.

More specifically, we compare with~\cite{DBLP:conf/iccv/OhBFS15} on head region, upper body region and their fusion (see Table.~\ref{tab:state-of-the-arts}).
In~\cite{DBLP:conf/iccv/OhBFS15}, the performance is achieved based on features extracted from fine-tuned CNNs without incorporating 
any contextual information. 
As shown in Table.~\ref{tab:state-of-the-arts},
our sequence prediction model outperforms~\cite{DBLP:conf/iccv/OhBFS15} by a reasonable margin in all of four settings. 
In the original setting, our result ($84.93\%$) also outperforms that of 
Zhang~\etal~\cite{DBLP:conf/cvpr/ZhangPTFB15} ($83.05\%$), although
we just use two body regions. 
In comparison, features from $107$ poselet regions are used in~\cite{DBLP:conf/cvpr/ZhangPTFB15}.

We are aware of that by using better features learnt from external data, it is possible to achieve higher recognition performance. 
For instance, higher accuracy on the day setting is reported in
~\cite{DBLP:conf/cvpr/LiBLSH16}, in which the features are extracted from the face region using a CNN trained for the face recognition task.
However, in our work, we just use features fine-tuned on the PIPA training set. 
We are interested to show how our sequence prediction formulation models the contextual information (including the relation context and scene context) and visual cues in a unified framework for person recognition, which 
improves the recognition performance.   

\subsection{Visualization}
\label{subsec:visualization}

We provide some visual examples of predicted instances in the PIPA test set by our sequential model in Fig.~\ref{fig:examples}. 
As shown in Fig.~\ref{fig:examples}, in most photos, our 
model correctly recognize the identities in the photo, including 
some challenging cases, such as non-frontal faces. 

%% file: conclusion.tex
In this work, we have introduced a sequence prediction formulation for the task of person recognition in photo albums. 
The advantage of our approach is that it can model both the rich contextual information in the photo, and individual's appearance in a unified framework. 
We trained our model end-to-end and witnessed a significant 
boost in the recognition performance, compared with 
baselines and state-of-the-approaches which do not exploit the
contextual information for the task. 